\documentclass[conference]{IEEEtran}
\IEEEoverridecommandlockouts
\usepackage{cite}
\usepackage{amsmath,amssymb,amsfonts}
\usepackage{algorithmic}
\usepackage{graphicx}
\usepackage{textcomp}
\usepackage{xcolor}
\usepackage{soul}
\usepackage{algorithm}
\usepackage{algorithmic}
\usepackage{arydshln}
\usepackage{makecell}
\usepackage{svg}
\usepackage[left=0.65in, right=0.65in, bottom=1.05in, top=0.72in]{geometry}

\usepackage[caption=false]{subfig}
\def\BibTeX{{\rm B\kern-.05em{\sc i\kern-.025em b}\kern-.08em
    T\kern-.1667em\lower.7ex\hbox{E}\kern-.125emX}}

\renewcommand{\baselinestretch}{0.931}

\begin{document}

\title{A Meta-learning based Generalizable Indoor Localization Model using Channel State Information \thanks{This material is based upon work supported by the Air Force Office of Scientific Research under award number FA9550-20-1-0090 and the National Science Foundation under Grant Numbers  CNS-2204445 and CNS-2232048.} \thanks{DISTRIBUTION STATEMENT A: Approved for Public Release; distribution unlimited AFRL-2023-0511 on 6 June 2023.}
}

\author{
	\IEEEauthorblockN{
	Ali Owfi\IEEEauthorrefmark{1}, 
        ChunChih Lin\IEEEauthorrefmark{1},
        Linke Guo\IEEEauthorrefmark{1},
        Fatemeh Afghah\IEEEauthorrefmark{1},
        Jonathan Ashdown \IEEEauthorrefmark{2},
        Kurt Turck \IEEEauthorrefmark{2}}

	\IEEEauthorblockA{\IEEEauthorrefmark{1}Holcombe Department of Electrical and Computer Engineering, Clemson University, Clemson, SC, USA \\
Emails: \{aowfi, 
chunchi, 
linkeg, 
fafghah\}@clemson.edu}
\IEEEauthorblockA{\IEEEauthorrefmark{2}Air Force Research Laboratory,  Rome, NY 13441, USA \\
	 Emails: \{jonathan.ashdown,kurt.turck\}@us.af.mil}
}



\maketitle

\begin{abstract}
Indoor localization has gained significant attention in recent years due to its various applications in smart homes, industrial automation, and healthcare, especially since more people rely on their wireless devices for location-based services. Deep learning-based solutions have shown promising results in accurately estimating the position of wireless devices in indoor environments using wireless parameters such as Channel State Information (CSI) and Received Signal Strength Indicator (RSSI). However, despite the success of deep learning-based approaches in achieving high localization accuracy, these models suffer from a lack of generalizability and can not be readily-deployed to new environments or operate in dynamic environments without retraining. In this paper, we propose meta-learning-based localization models to address the lack of generalizability that persists in conventionally trained DL-based localization models. Furthermore, since meta-learning algorithms require diverse datasets from several different scenarios, which can be hard to collect in the context of localization, we design and propose a new meta-learning algorithm, TB-MAML (Task Biased Model Agnostic Meta Learning), intended to further improve generalizability when the dataset is limited. Lastly, we evaluate the performance of TB-MAML-based localization against conventionally trained localization models and localization done using other meta-learning algorithms. 

\end{abstract}

\begin{IEEEkeywords}
Wireless Indoor Localization, Channel State Information (CSI), Meta-Learning 
\end{IEEEkeywords}

\section{Introduction}


Contrary to outdoor localization, where line-of-sight (LOS) is present in most instances, there are a lot of challenges in indoor localization, such as the presence of physical barriers, multipath effect, and the complexity of indoor environments. 
These challenges have been widely studied, and through recent works, data-driven and Machine Learning (ML) approaches have shown promising results for indoor localization \cite{singh2021machine}. 
Traditional indoor localization methods, such as geometric-based approaches (multilateration, trilateration, and triangulation) or fingerprinting, rely on manual calibration, which can be time-consuming and labor-intensive. Moreover, these methods tend to be less accurate than data-driven methods, especially in complex indoor environments with obstacles and signal interference.

Many technologies have been studied as a medium for wireless-based indoor localization, such as ultrasonic, radio frequency identification (RFID),  ultra-wideband (UWB), Bluetooth, and WiFi\cite{li2019machine}. Out of the proposed technologies, Wi-Fi is often preferred for indoor localization due to its widespread availability, low cost, and ease of implementation. Moreover, Wi-Fi signals also have a relatively large coverage area, meaning fewer access points are needed to cover a given indoor space. 


Most proposed Wi-Fi-based indoor localization models either use Received Signal Strength Indicator (RSSI) or Channel State Information (CSI), as both these parameters can provide valuable information regarding the location of wireless devices. RSSI is simple and very easy to obtain as it does not require any special hardware to acquire. However, it is very volatile, and its information is coarse because RSSI is simply the strength of the received wireless signal. In contrast, CSI provides information about the channel characteristics between a device and an access point. CSI can provide more detailed information about the wireless signal, including phase and amplitude in different sub-channels. Although CSI is more stable than RSSI, it is also volatile and susceptible to any environmental changes.

Even though these parameters are not perfect, many data-driven localization models have been proposed that incorporate one of the mentioned parameters, a mixture of them, or a processed version of them in their training dataset, and perform relatively well on the respective testing dataset\cite{nessa2020survey}. The issue with these models is that they have been trained on a train-set collected from a specific location and at a specific time, and due to the high volatility of the mentioned parameters, the underlying distribution that the data-driven model has learned from the given train-set is certain to change when the environment changes or even with time. This means that the learned information for a specific location and time is nearly ineffective for other locations or the same location at a different time. For these conventionally trained ML models to perform well in new environments, they have to go through a complete process of training, which makes these models not be readily-deployable for new locations. Moreover, a complete training process can be very hard or even not feasible in some instances due to the limitations on resources, time, and new datasets. All these mentioned reasons render conventionally trained ML models impractical as a scalable solution for indoor localization.

This paper aims to solve the aforementioned issues with conventionally trained indoor localization models. We propose a generalizable indoor localization model using meta-learning, which can utilize the knowledge gained from training on multiple datasets collected in different environments towards new unseen environments requiring very little fine-tuning. 
To this end, we have collected CSI data in 33 different locations, with the data in each location constituting a separate task. We then evaluate the generalizability of the proposed meta-learning-based localization model and other benchmark methods by training on a set of the collected tasks and testing against a set of unseen tasks. Meta-learning algorithms require a sizeable amount of training tasks, which is time-consuming and challenging to collect in the context of indoor localization. Thus, we propose a data-efficient novel meta-learning algorithm, Task Biased Agnostic Meta Learning (TB-MAML), based on Model Agnostic Meta Learning (MAML) \cite{finn2017model} to further improve generalizability even with relatively limited datasets. Lastly, we compare the generalizability of the TB-MAML-based localization model with other meta-learning-based localization models in terms of the number of tasks used for training.

The rest of this paper is structured as follows: Section II discusses the previous works on wireless indoor localization. Section III gives a brief introduction to meta-learning and MAML, followed by a description of our proposed meta-learning algorithm and the overall design of our indoor localization model. Section IV describes the dataset we have collected, explains the experiments we used to evaluate our proposed model, provides the evaluations, and discusses them. Finally, section V concludes the paper.
\section{Related Work}

Many of the earlier works focused on using Received Signal Strength Indicator (RSSI) as a measurement to determine the location of wireless devices \cite{sugano2006indoor,zhu2013rssi}. In \cite{bahl2000radar}, RSSI values for multiple reference points within an indoor perimeter are measured and stored. In the online phase, the RSSI values from three indoor APs are compared against the stored RSSI dataset based on the Euclidean distance. A weighted average is then calculated using the similarity of the new RSSI readings and the stored reference points. 
Horus\cite{youssef2005horus} is another localization scheme that employs a probabilistic approach and utilizes RSSI data. In Horus, location-clustering
techniques are implemented to reduce the computational requirements of
the algorithm. 
In \cite{alhajri2019indoor}, the authors built a two-stage localization system based on K-Nearest Neighbors (KNN). In the first stage, their algorithm aims to identify the type of environment, and in the second stage, localization is performed using KNN.
They utilized RSSI alongside a hybrid feature vector of Channel Transfer Function (CTF) and Frequency Coherence Function (FCF). They concluded that a model using multiple or hybrid features outperforms RSSI-only approaches. 

While RSSI is simple and very easy to obtain, the information it carries about the channel is coarse as it only has one signal strength reading for each packet. As an alternative and a more reliable source of information, Channel State Information (CSI) can be used for localization \cite{xiao2012fifs}. CSI measures the amplitude and phase of the received signal at each subcarrier, providing detailed information about the channel characteristics.

DeepFi \cite{wang2015deepfi} proposes a Deep Neural Networks (DNN) model for indoor localization that uses CSI amplitude for its input. A greedy learning algorithm is used to train the model to reduce the computational complexity. Finally, in the online localization phase, DeepFi uses a probabilistic method based on the radial basis function to estimate the target's location. Evaluations indicate that DeepFi outperforms traditional statistical localization schemes such as HORUS\cite{youssef2005horus} and FIFS\cite{xiao2012fifs}.

ConFi \cite{chen2017confi} is the first localization paper that utilizes Convolutional Neural Networks (CNN). As CNNs are powerful tools for inferring information from images, ConFi arranges CSI amplitude data to create CSI feature images. The created feature images are then fed to CNN with three convolutional and two fully connected layers. ConFi treats localization as a classification problem ,where inputs are localized based on several specified reference points. Their evaluations show that ConFi outperforms other conventional data-driven localization methods, demonstrating that CNN-based localization is a viable option. 

In CiFi\cite{wang2018cifi}, CSI phase data was used as a medium to calculate the angle of arrival (AoA). They used the Intel 5300 network interface card with three antennas to collect the CSI data. Based on the measured CSI phase data for every two adjacent antennas, the phase difference was obtained, from which AoA can be calculated. As AoA is not as random raw CSI phase data, it was then fed to the CNN-based localization model they proposed as an input. Their results show that CiFi can compete with other established localization methods, such as DeepFi, suggesting that CSI phase data can also be effective for localization.

One fundamental issue with most of the mentioned localization models is the lack of generalizability and adaptability to new or dynamic environments, as these models have to be retrained when the environment changes to perform well. This dramatically hinders their applicability to real-world scenarios. To address this issue, a few recent works have utilized transfer learning and domain adaptation.

Transloc \cite{li2021transloc}, is a knowledge transfer framework for indoor localization, which derives a cross-domain mapping to
transfer the specific knowledge of one domain to another and then creates a homogeneous feature space. This enables the localization model to perform well when the environment changes with a limited number of new training data from the new environment. To increase robustness against environmental changes, Fidora \cite{chen2022fidora} augments the data with a variational autoencoder to add diversity and then employs a domain-adaptive classifier to adjust the localization model to the new data.

In a recently published work, authors of \cite{gao2022metaloc} utilized meta-learning for indoor localization to increase the generalizability of DL-based localization models. \cite{gao2022metaloc} proposes a localization framework based on  MAML \cite{finn2017model} as opposed to conventional DL-based localization models. The results presented in this paper are based on simulated RSSI data. Some parameters used to generate the simulated data, such as the room size, number of reference points, and noise level, differed for each scenario, the parameters being set by pre-determined settings for training and testing scenarios separately. This was done to increase the diversity of scenarios. As RSSI is highly dependent on many parameters, such as obstacles, obstructions, and positioning,  which simulations can not fully capture. Hence, the generated scenarios may not be realistically diverse. In the context of meta-learning, a lack of sufficient diverse training scenarios may lead to meta-overfitting in the model, memorizing the learning process for a handful of scenarios and not reaching generalizability for unseen scenarios.



\section{Methodology}

\subsection{Meta-Learning} 
Meta-learning, also known as "learning to learn," is a subfield of machine learning that focuses on developing algorithms capable of quickly adapting to new tasks with limited data. In conventional deep learning, models are designed to perform well on a specific task with a fixed objective over a dataset divided into training and testing sets. However, meta-learning aims to improve a learning model's performance by training it to learn the learning process itself, enabling it to adapt to new tasks with a very small dataset and consequently in a shorter amount of time. In meta-learning, multiple tasks are divided into training tasks and testing tasks, and each task consists of a distinct objective, a support set (training set), and a query set (test set). In the inner loop (also referred to as the adaptation phase), the meta-learning model adapts to each task by training on the corresponding support set, followed by computing the loss function for that task on the query set. It should be noted that the outer objective function utilized in meta-learning models for overall learning is not the same as the objective function used for each task during the inner loop.

\subsection{Model-Agnostic Meta-Learning (MAML)} \label{sec:MAML}

Among the many proposed meta-learning algorithms, MAML \cite{finn2017model} is arguably the most popular algorithm. One reason for this popularity is that, as its name suggests, MAML is model agnostic, meaning that it can be applied to any differentiable model regardless of its architecture or specific learning objective. MAML aims to determine an initial set of parameters for the inner model, such that adapting to new tasks can be done as quickly as possible using the computed initial set of parameters. Formally, MAML considers an inner model $f$ with a set of parameters $\theta$ denoted by $f_\theta$.

During the inner loop, for each task $\mathcal{T}_i$, the model adapts to task $\mathcal{T}_i$ by training on the corresponding support set and, respectively, updating model parameters $\theta$ based on the inner objective function to compute $\theta'_i$. The following equation shows the adaptation phase of a single gradient step, but it can be extended to cases where multiple gradient steps are taken, as well.

\begin{equation}
    \theta_{i}^{'} = \theta - \alpha\Delta_{\theta}\mathcal{L}_{\mathcal{T}_i}(f_{\theta})
\end{equation}
where $\alpha$ is the step size.

The outer objective function used in the outer loop is defined as below:
\begin{equation}
\min _{\theta} \sum_{\mathcal{T}_{i} \sim p(\mathcal{T})} \mathcal{L}_{\mathcal{T}_{i}}\left(f_{\theta_{i}^{\prime}}\right)=\sum_{\mathcal{T}_{i} \sim p(\mathcal{T})} \mathcal{L}_{\mathcal{T}_{i}}\left(f_{\theta-\alpha \nabla_{\theta}} \mathcal{L}_{\mathcal{T}_{i}}\left(f_{\theta}\right)\right)
\end{equation}
where $f_\theta'$ is optimized with respect to  the initial set of parameters $\theta$ used to adapt to each task. And the outer loop optimization rule is as followings:
\begin{equation}
\theta \leftarrow \theta-\beta \nabla_{\theta} \sum_{\mathcal{T}_{i} \sim p(\mathcal{T})} \mathcal{L}_{\mathcal{T}_{i}}\left(f_{\theta_{i}^{\prime}}\right)
\label{eq:outer_update}
\end{equation}
where $\beta$ is a hyper-parameter known as \emph{meta-step size}.

For all training tasks, the inner loop is performed, and then $\theta$ is updated during the outer loop as shown in (\ref{eq:outer_update}). In contrast, only the inner loop is performed for the testing tasks to see how well the model can adapt to an unseen task using a limited support set.

\subsection{Task Biased Model Agnostic Meta Learning (TB-MAML)}

\begin{figure}[htp]
\begin{center} 
\includegraphics[width=0.4\textwidth]{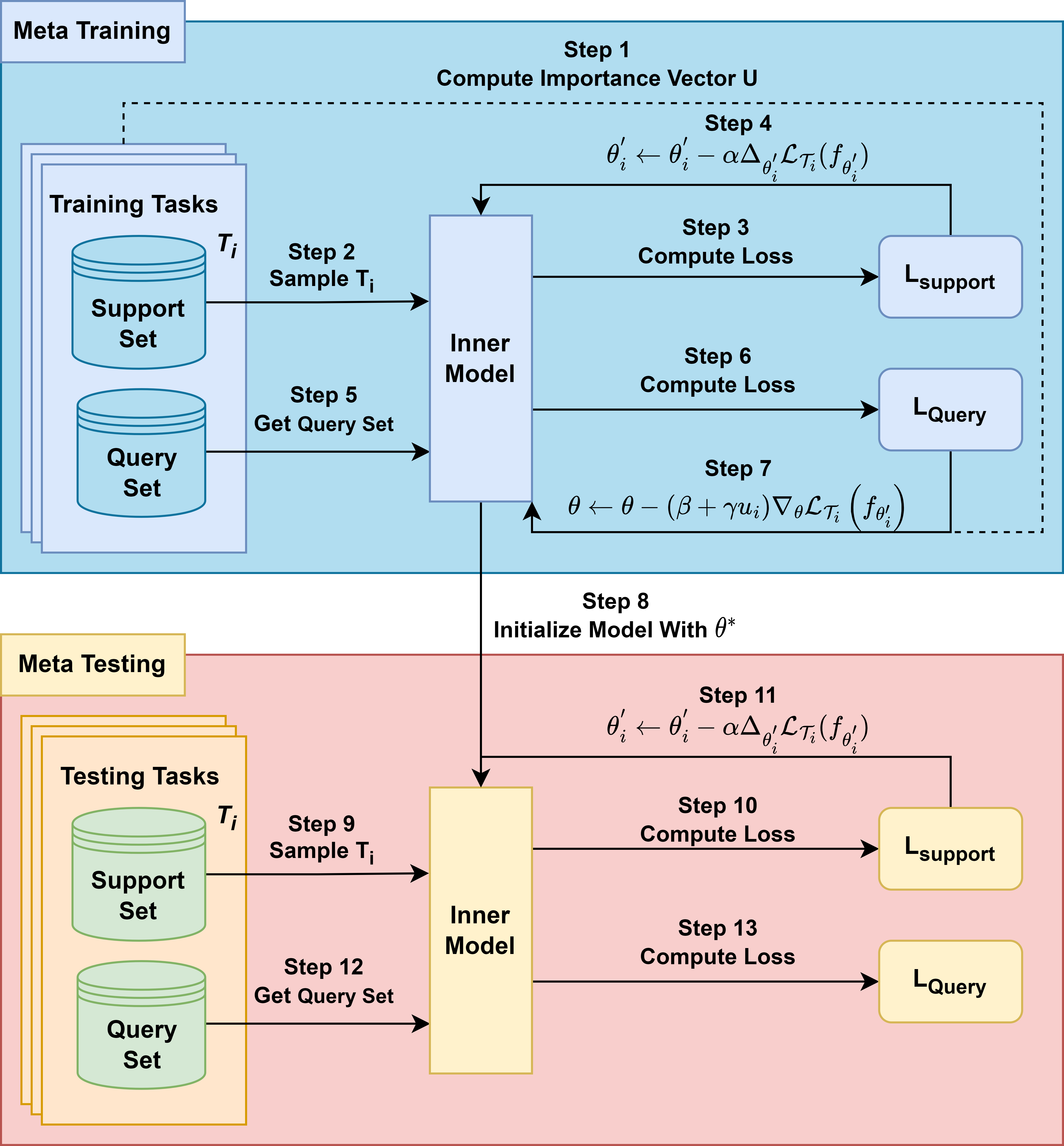}
\caption{Schematic of the proposed TB-MAML algorithm.} 
\label{fif:adamaml_scehmatic} 
\end{center} 
\end{figure}

In this section, we would like to propose TB-MAML, a novel meta-learning algorithm based on MAML. TB-MAML is designed for cases with a  limited number of training tasks for the meta-training process. In conventional deep learning, not having enough data samples leads to overfitting, memorization of the data samples, and consequently, not learning the underlying distribution from which the data was sampled. A similar concept called \textit{Meta-overfitting} exists in the context of meta-learning. 
Consider a distribution over all tasks $\mathcal{P(T)}$ and a limited set of tasks $\mathcal{T} $ that do not wholly represent the distribution $\mathcal{P(T)}$. Suppose a meta-learning model just uses the tasks $\mathcal{T} $ for the meta-training process. In that case, it will meta-overfit to the tasks in $\mathcal{T} $, meaning that it will not learn to adapt quickly to all the tasks drawn from the distribution $\mathcal{P(T)}$, but just the tasks in $\mathcal{T} $. TB-MAML is designed to learn the underlying distribution $\mathcal{P(T)}$ even in cases where the set of training tasks $\mathcal{T} $ available to us is limited. 
In the context of localization, each task requires a training set and a test set for multiple reference points in a location. Since the process of collecting data for multiple reference points per each task is time-consuming, gathering a large enough number of indoor localization tasks is not an easy feat.
To provide a sense of comparison, the dataset Omniglot which is a standard toy dataset for meta-learning literature has 1623 classes. If we define each task as a 10-way classification, we will have $1623 \choose 10$ different tasks at our disposal which we can split into meta-training and meta-testing tasks. 
To this end, TB-MAML is particularly valuable in the context of indoor localization as it is designed for improved generalizability for circumstances where the number of tasks is limited.

TB-MAML defines an importance vector over the available meta-training tasks to identify the tasks that push the model more toward generalizability, or in other words, the tasks that provide better information regarding the learning process of all the other tasks in $\mathcal{P(T)}$. TB-MAML is biased towards the more important tasks as it emphasizes them during the learning process, hence the name, Task Biased Model Agnostic Meta Learning.

To calculate the importance vector, we first select task $i$ from the meta-training tasks. We train our inner model with the training set of task $i$. In a case of $n$-shot learning, for each task $j$ in the meta-training tasks where $i \neq j$, we further train the inner model with the support set of task $j$ and then test the model against the query set of task $j$, resulting in the loss $\mathcal{L}_{i}({\theta}_{ij})$. We denote the average of all these losses as $\mathcal{L}_{i}$, which is a measurement of how well a model trained for task $i$ can adapt to unseen tasks. By calculating the average loss $\mathcal{L}_{i}$ for all tasks, we form the vector $[\mathcal{L}_{1},...,\mathcal{L}_{n}]$. By normalizing this vector between values (-1,1) and then inverting the values, we derive the importance vector $[{u}_{1},...,{u}_{n}]$.

During outer loop (steps 6 and 7 in fig \ref{fif:adamaml_scehmatic}), when the inner loop TB-MAML has adapted to the task $j$ using the corresponding support set, it updates $\theta$ based on the importance of task $j$. More Formally:

\begin{equation}
\theta \leftarrow \theta-(\beta + \gamma u_j) \nabla_{\theta} \sum_{\mathcal{T}_{i} \sim p(\mathcal{T})} \mathcal{L}_{\mathcal{T}_{i}}\left(f_{\theta_{i}^{\prime}}\right)
\end{equation}
where $u_j$ is the importance of task $j$ and $\gamma$ is a hyperparameter that adjusts intensity of the importance vector.

The entire process of TB-MAML is summarized in Algorithm \ref{alg:TB-MAML}. Furthermore, a schematic of TB-MAML is provided in Fig \ref{fif:adamaml_scehmatic} for illustration of TB-MAML. In step 1, the importance vector is computed from the training tasks available. In step 2, the inner model is initilized with weight $\theta$, task $\mathcal{T}_{i}$ is sampled and the corresponding support set is fed to the inner model. Steps 3 and 4 represent the inner loop where the model adapts to task $\mathcal{T}_{i}$. In step 5, query set of $\mathcal{T}_{i}$ is given to the model and outer loop is then performed (steps 6 and 7), and the inner model's initialization weight $\theta$ is updated. After sufficient iterations when convergence is reached, meta-testing phase starts (steps 9-13). The steps taken in this phase are similar to the ones taken in the meta training phase, with the difference that outer loop is not performed.

\begin{algorithm}
\caption{TB-MAML}
\begin{algorithmic} 
\REQUIRE $\mathcal{P(T)}$: Distribution over training tasks
\REQUIRE $U=[{u}_{1},...,{u}_{n}]$: Importance vector for training tasks
\REQUIRE $\alpha, \beta, \gamma,$: inner step size, outer step size, and importance vector intensity

\STATE Randomly initialize inner model's weights $\theta$
\WHILE{not converged}

\STATE Sample meta-training task $T_i \sim \mathcal{P(T)}$ 
\STATE $\theta_{i}^{'} \gets \theta$
\FORALL{inner loop iterations}
\STATE Using support set $\mathcal{D}_{i}$ compute loss $\mathcal{L}_{\mathcal{T}_{i}}$
\STATE Update $\theta_{i}^{'} \leftarrow \theta_{i}^{'} - \alpha\Delta_{\theta_{i}^{'}}\mathcal{L}_{\mathcal{T}_i}(f_{\theta_{i}^{'}})$
\ENDFOR

\STATE Using query set $\mathcal{D}_{i}^{\prime}$ compute loss $\mathcal{L}_{\mathcal{T}_{i}}$
\STATE Update $\theta \leftarrow \theta-(\beta + \gamma u_i) \nabla_{\theta}\mathcal{L}_{\mathcal{T}_{i}}\left(f_{\theta_{i}^{\prime}}\right)$

\ENDWHILE

\end{algorithmic}
\label{alg:TB-MAML}
\end{algorithm}



\section{Evaluations}

\subsection{Dataset}

\begin{figure}[htp]
\centering
\subfloat[Reference points arrangement]{%
  \includegraphics[clip,width=0.7\columnwidth]{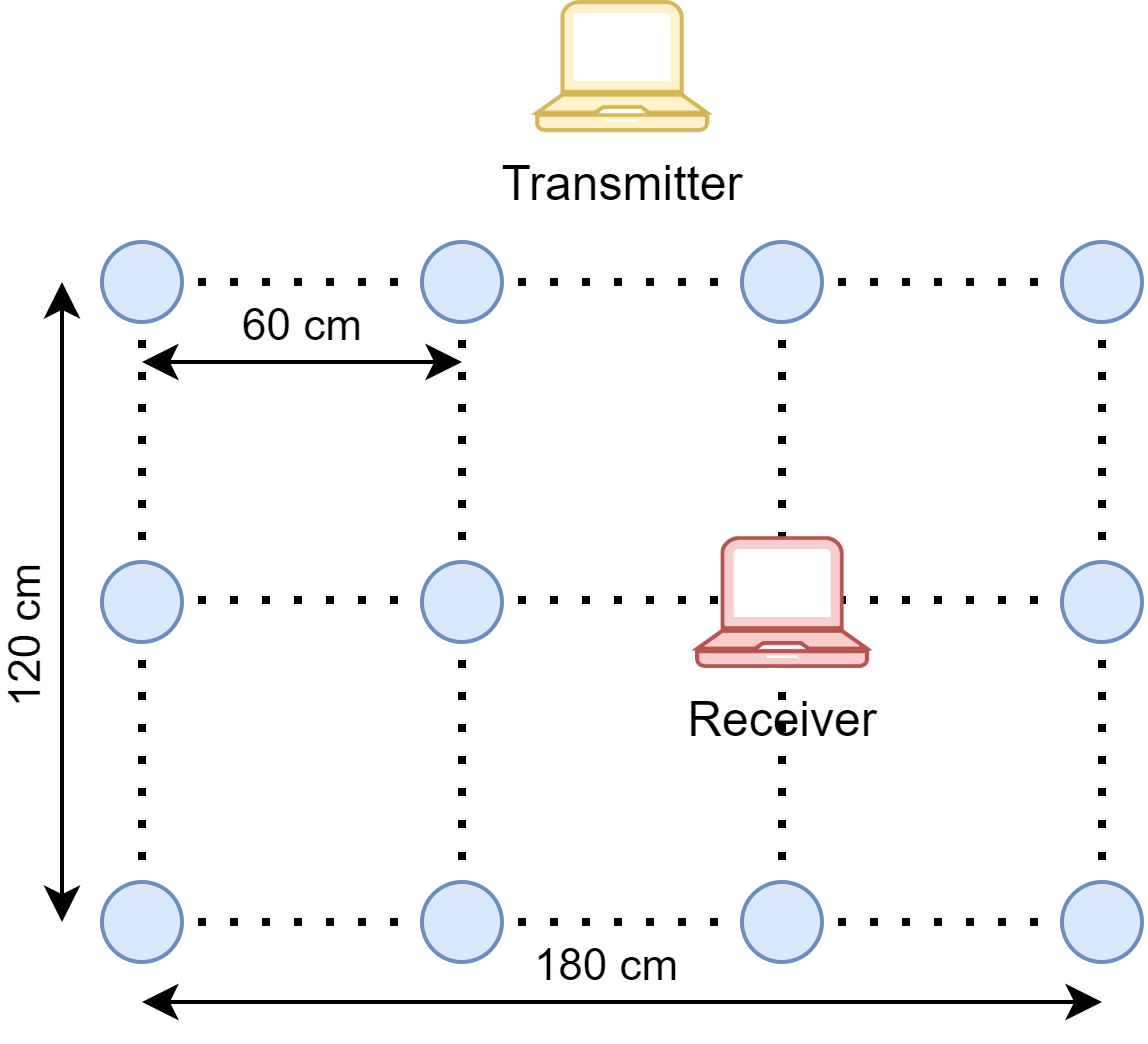}%
}

\subfloat[Example locations of different scenarios]{%
  \includegraphics[clip,width=0.98\columnwidth]{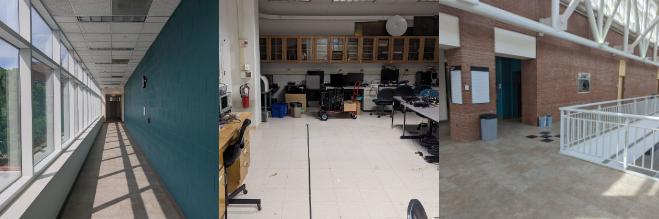}%
}

\caption{Experiment Settings.}
\label{fig:RP_positions}
\end{figure}

For the purpose of testing the generalizability and adaptability of the discussed localization models, a dataset consisting of multiple different scenarios was required. In total, we collected 33 scenarios, each scenario resulting in a different task. All 33 scenarios were collected in different indoor locations such as rooms, laboratories, corridors, and auditoriums, to diversify the overall dataset as much as possible. A few example locations can be seen in fig \ref{fig:RP_positions}(b).
Each scenario consisted of 12 reference points, arranged in a 3 by 4 grid with a grid size of 60 cm. Fig \ref{fig:RP_positions}(a) shows the positioning of the reference points in test scenarios. We collected CSI data for all reference points using two Intel 5300 network interface cards, one as a receiver and one as a transmitter. We transmitted Wi-Fi 802.11n packets with 20 MHz bandwidth on the 5 GHz frequency band and for every reference point. The transmitter transmits 40 bursts each of the burst includes 100 packets. To counter the instantaneous interference or fluctuations in the environment, each burst has 1 second pause time before the next one. The transmitter uses only one antenna for transmission, while the receiver uses all three antennas for receiving. In 802.11n, 52 subcarriers are carrying information and used for calculating the CSI data. The Intel 5300 card follows a grouping method that reduces the size of the CSI report field to 30. Hence, each CSI sample had a size of $3\times 30$. We calculate and normalize only the amplitude of the CSI data before feeding it into the network. 

\setlength{\tabcolsep}{6pt} 
\renewcommand{\arraystretch}{1} 
\begin{table}[]

\centering
\caption{Structure of Inner Model}
\begin{tabular}{cccc}
\Xhline{3\arrayrulewidth} 
\multicolumn{1}{c}{\textbf{Layer}} & \multicolumn{1}{c}{\textbf{Input}} & \multicolumn{1}{c}{\textbf{Parameters}}                                            & \multicolumn{1}{c}{\textbf{Activation Function}} \\ \Xhline{3\arrayrulewidth} \vspace{4pt}
1D Convolution                       & 3*30                                & \begin{tabular}[c]{@{}c@{}}Out Channels=10\\ Kernel Size=3\\ Padding=1\end{tabular} & ReLU                                              \\ \vspace{4pt}
1D Max Pooling                       & 10*30                               & Kernel Size=2                                                                       & -                                                 \\ \vspace{4pt}
1D Convolution                       & 10*15                               & \begin{tabular}[c]{@{}c@{}}Out Channels=15\\ Kernel Size=3\\ Padding=1\end{tabular} & ReLU                                              \\ \vspace{4pt}
1D Max Pooling                       & 15*15                               & Kernel Size=2                                                                       & -                                                 \\ \vspace{4pt}
Dense                                & 105                                 & 128 neurons                                                                         & ReLU                                              \\ \vspace{4pt}
Dense                                & 128                                 & 64 neurons                                                                          & ReLU                                              \\ \vspace{4pt}
Dense                                & 64                                  & 32 neurons                                                                          & ReLU                                              \\ \vspace{4pt}
Dense                                & 32                                  & 8 neurons                                                                           & ReLU                                              \\ \vspace{4pt}
Dense                                & 8                                  & 2 neurons                                                                           & -                                              \\ 


\\ \Xhline{3\arrayrulewidth}
\end{tabular}

\label{tab:inner_model}
\end{table}





\begin{figure}[htp]
\centering



  \subfloat[][No additional training]{\includegraphics[width=0.45\columnwidth]{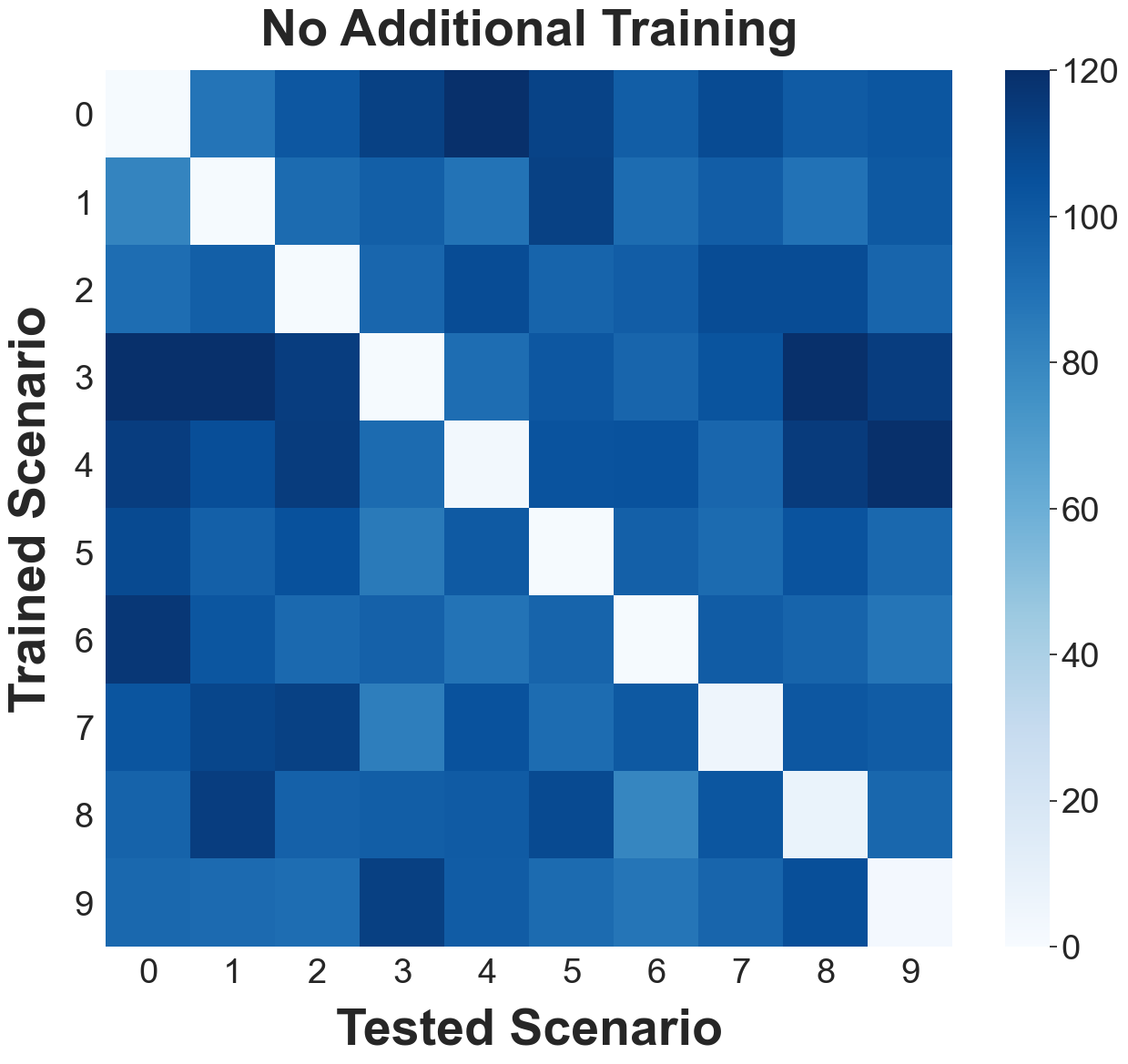}}
\hspace{0.02\columnwidth}
 \subfloat[][Additional 5 shot training]{\includegraphics[width=0.45\columnwidth]{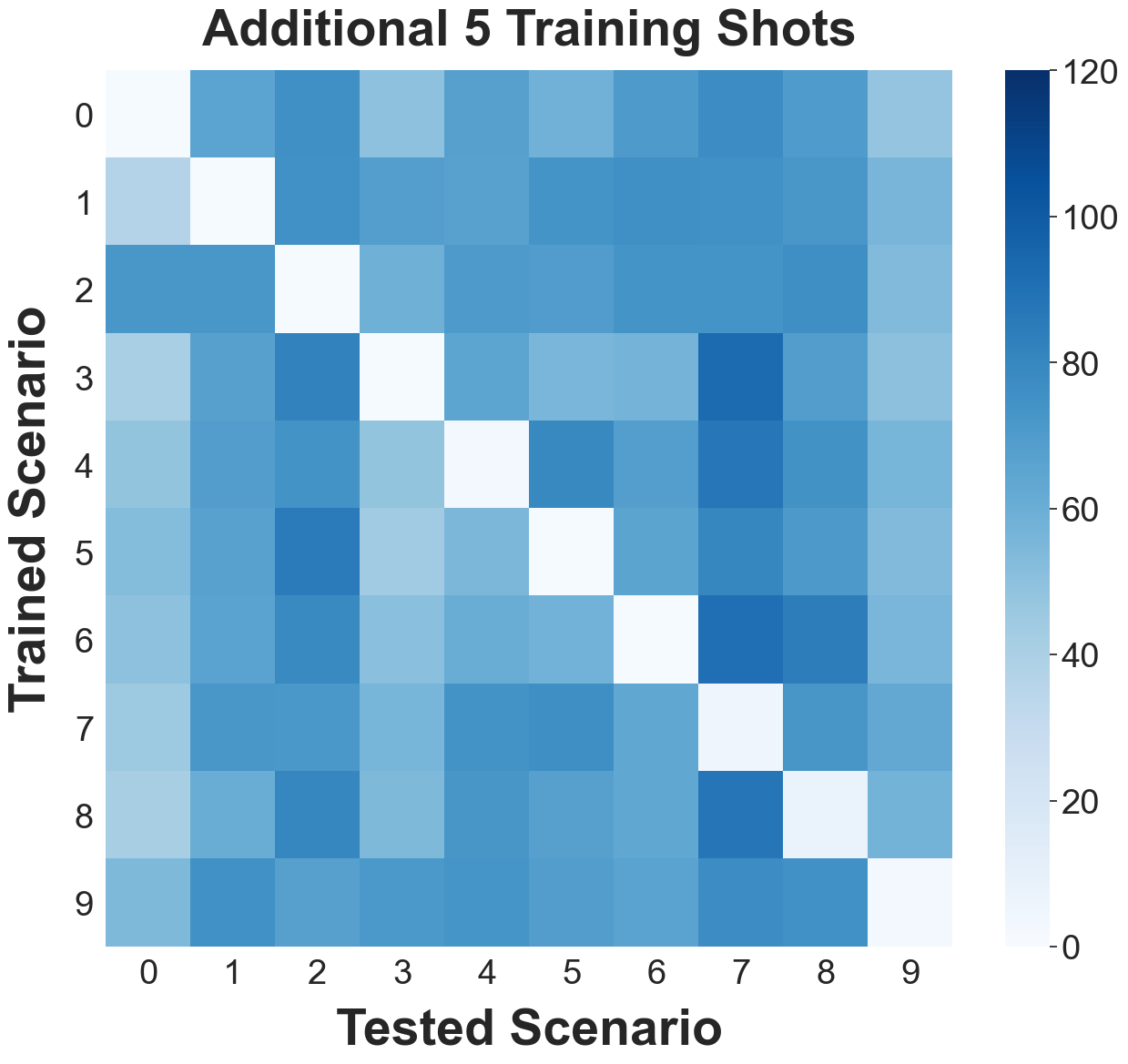}}

\caption{Localization distance errors of a conventional DL-based localization model trained on scenario $i$ and tested against scenario $j$. In Figure (a) no additional training samples from the testing scenario were provided to the model, whereas in Figure (b), five additional samples per reference point from the testing scenario were given to the model for further training. For cleaner visualization purposes, only the first 10 scenarios are considered in the figures.}
\label{fig:heatmap}
\end{figure}

\subsection{Generalizability Analysis of Conventional DL-based Localization Models}
Before providing the results for the proposed meta-learning models we would like to emphasize on the lack of generalizability in conventional DL localization models. Fig \ref{fig:heatmap}(a) depicts the error of a conventionally trained DL localization model on one task and then tested against another one. The architecture of the used DL model is described in table \ref{tab:inner_model}. The value in cell $(i,j)$ is the distance error of the localization model trained for scenario $i$ and then tested against scenario $j$. For cleaner visualization purposes, only the first 10 scenarios are considered in the heatmap. As it can be seen from the figure, the distance error on the main diagonal is very low (when the model was trained for scenario $i$ and was tested against $i$) but for the other cases we can see the distance error is pretty high, pointing to the lack of generalizability of the conventionally trained localization model. The mean distance error in this plot is 95.98 $cm$.

In \ref{fig:heatmap}(b), we have the same experiment as \ref{fig:heatmap}(a) but just 5 new data samples per reference point from scenario $j$ are provided to the localization model to train on. With a mean distance error of 63.45cm, we can observe that the overall distance error has reduced as expected in comparison with \ref{fig:heatmap}(a). But the distance error is still very high when compared to the main diagonal of the heatmap, pointing to the lack of adaptability in the conventionally trained model.

\subsection{Localization Accuracy Analysis}

To evaluate the generalizability of our proposed TB-MAML-based localization model, we are considering several benchmark algorithms in our experiments. The first benchmark, referred to as conventional learning, we have a localization model without prior training that has to train on a few new samples from the unseen environments. In Transfer Learning, we are feeding the full training dataset of one of the scenarios to the localization model, followed by a few new samples from the unseen target environments. We are then employing MAML, First Order Model Agnostic Meta Learning (FOMAML), and our proposed meta-learning model, TB-MAML, as cases of meta-learning based localization. It has to be noted that for all benchmarks, results are based on localization accuracies from unseen scenarios. All algorithms have been executed multiple times with different training scenarios and testing scenarios and results are averaged over the runs to reduce randomness in results. To have a fair comparison, the same inner model structure has been used for all cases which is described in table \ref{tab:inner_model}.

\begin{figure}[htp]
\centering


\subfloat[][5 Shot]{\includegraphics[width=0.45\columnwidth]{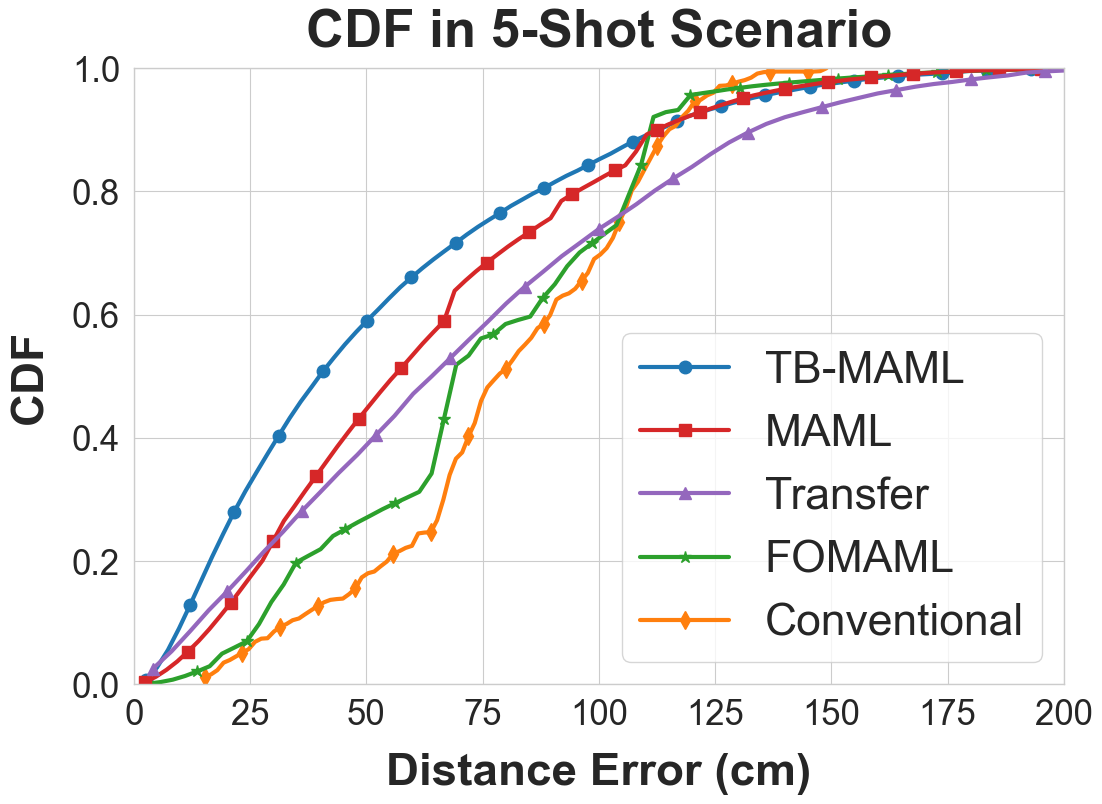}}
\hspace{0.02\columnwidth}
 \subfloat[][3 Shot]{\includegraphics[width=0.45\columnwidth]{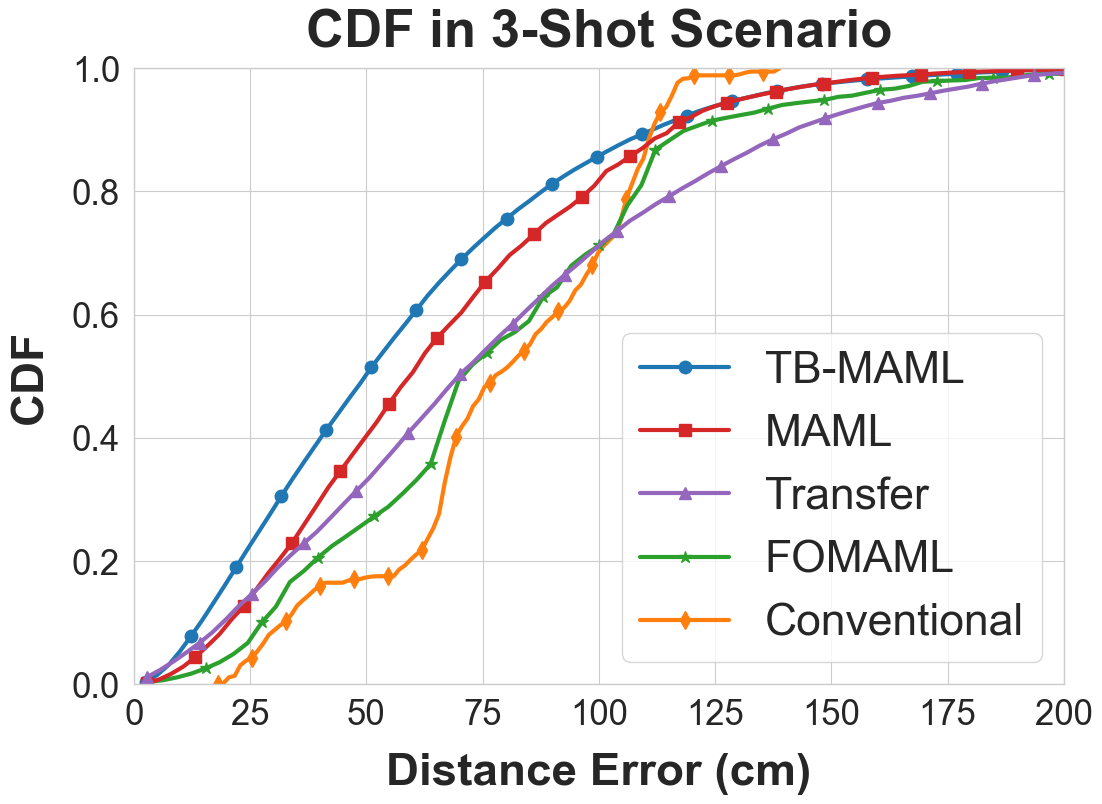}}

\caption{CDF of localization distance errors for different localization models. Figure (a) and (b) depict cases of 5-shot and 3-shot learning respectively. }
\label{fig:cdf}
\end{figure}

\begin{figure}[htp]
\centering


\subfloat[][5 Shot]{\includegraphics[width=0.49\columnwidth]{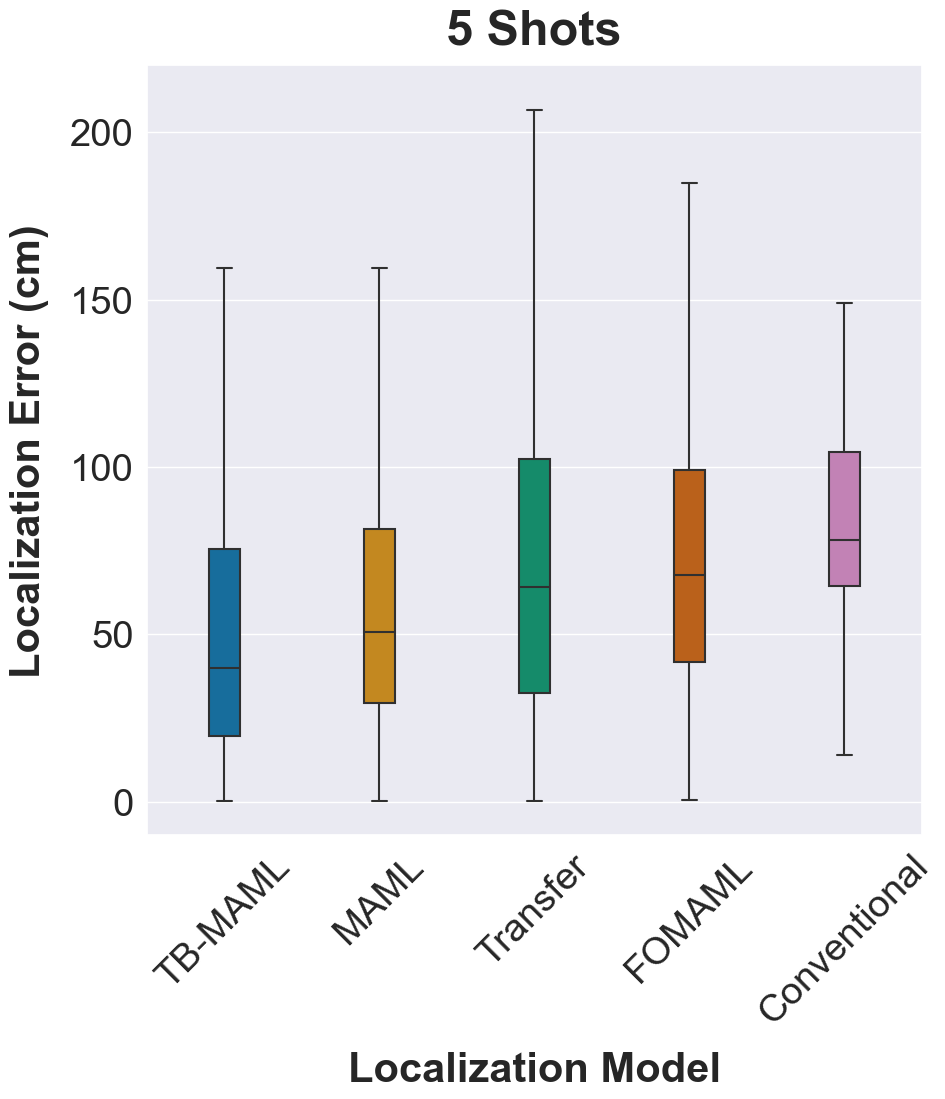}}
 \subfloat[][3 Shot]{\includegraphics[width=0.49\columnwidth]{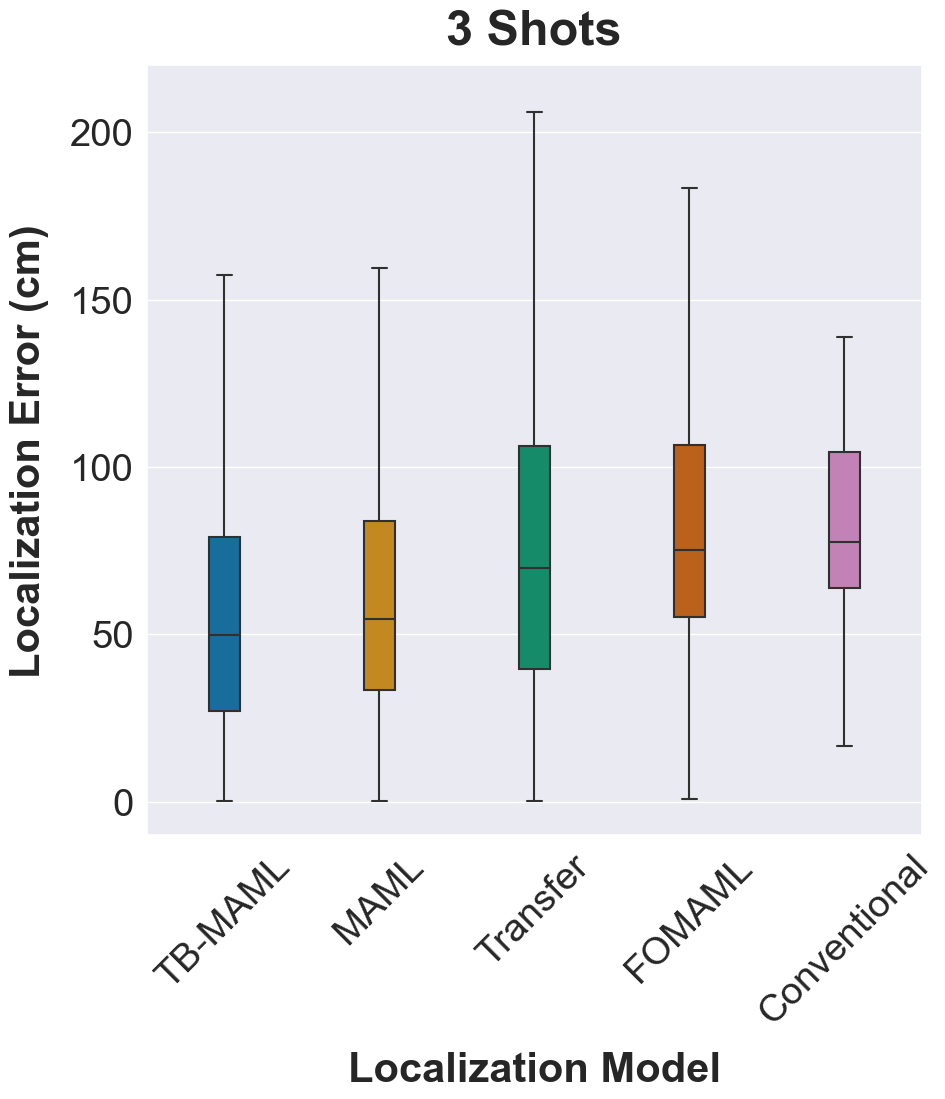}}

\caption{Distribution of localization distance errors for different localization models. Figure (a) and (b) depict cases of 5-shot and 3-shot learning respectively.}
\label{fig:boxplot}
\end{figure}

Figure \ref{fig:cdf} shows the localization errors of the compared localization models in terms of cumulative distribution function (CDF), in multiple cases with different number of new samples from the new scenarios. As visible from the figures, TB-MAML localization outperforms other benchmarks throughout all few-shot scenarios, followed by MAML. We can further observe that FOMAML-based localization is more accurate that a conventionally trained model, but slightly less accurate than transfer-learning-based localization. Since FOMAML is a computationally efficient first-order approximation of MAML and, therefore, a less accurate meta-learning algorithm, this observation is not unexpected. In the 5-shot case, 59 percent of distance errors for TB-MAML were below 50 cm, while the corresponding percentage for MAML, Transfer learning, FOMAML, and Conventional learning were 45, 38, 22, and 18 percent respectively. Figure \ref{fig:boxplot} depicts a boxplot of the distance errors for the same experiments. Again, it can be observed that TB-MAML localization outperforms other benchmarks in terms of the average distance error, followed by the MAML localization model. 

\subsection{Limited Number of Tasks Analysis}

\begin{figure}[htp]
\begin{center} 
\includegraphics[width=0.6\columnwidth]{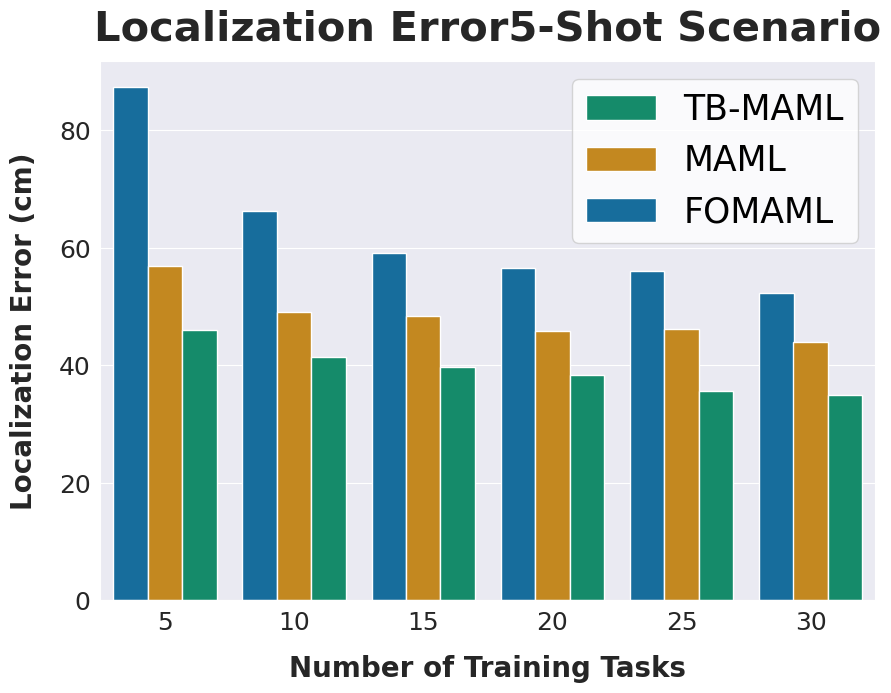}
\caption{Localization distance errors of meta-learning based localization models over the number of training tasks.} 
\label{fig:error_over_training_tasks} 
\end{center} 
\end{figure}

In another experiment, we compared the accuracy of the mentioned meta-learning based localization models with our proposed TB-MAML-based localization model in scenarios with different number of training tasks. Figure \ref{fig:error_over_training_tasks} illustrates the results for this experiment. As expected we can observe that distance error of all compared meta-learning-based algorithms increases as the number of training tasks decreases. But we can also observe that TB-MAML outperforms the other benchmark localization algorithms throughout all scenarios with different number of training tasks. Moreover, we can see that TB-MAML is less affected in comparison when the number of training tasks is small (e.g. five training tasks), as TB-MAML is designed for situations where the number training tasks is limited.

\section{Conclusion}

In this paper, we addressed the lack of generalizability of conventionally trained indoor localization models by proposing meta-learning-based localization. Moreover, we designed a new meta-learning algorithm, TB-MAML, specialized to reach better generalizability when the number of scenarios available for training a meta learning model is limited. This characteristic of TB-MAML is desired in the context of indoor localization, as collecting large enough diverse scenarios is difficult and time-consuming. Through extensive experimental results using real data collected from 33 different locations, we showed that meta-learning-based localization models dominate conventionally trained localization models in generalizability. Furthermore, in another experiment between meta-learning-based localization models, we showed that TB-MAML-based localization reaches better generalizability even in cases with extremely limited number of available training scenarios.

\bibliographystyle{IEEEtran}
\bibliography{main}

\end{document}